\documentclass[acmtog]{acmart}

\usepackage{graphicx} \graphicspath{{figures/}}
\usepackage{amsmath}
\usepackage[linesnumbered,ruled,vlined]{algorithm2e} 

\SetAlFnt{\small}
\SetAlCapFnt{\small}
\SetAlCapNameFnt{\small}
\SetAlCapHSkip{0pt}
\usepackage{acronym}
\usepackage{enumitem}
\usepackage{balance}
\usepackage{xspace}
\usepackage{setspace}
\usepackage[skip=3pt,font=small]{subcaption}
\usepackage[skip=3pt,font=small]{caption}
\usepackage[dvipsnames,svgnames,x11names,table]{xcolor}
\usepackage[capitalise,noabbrev,nameinlink]{cleveref}
\usepackage{booktabs,tabularx,colortbl,multirow,multicol,array,makecell,tabularray}
\usepackage{overpic,wrapfig}
\usepackage[misc]{ifsym}
\usepackage{pifont}

\usepackage{nicefrac}
\usepackage{bm}
\usepackage{dsfont}

\makeatletter
\DeclareRobustCommand\onedot{\futurelet\@let@token\@onedot}
\def\@onedot{\ifx\@let@token.\else.\null\fi\xspace}
\def\eg{\emph{e.g}\onedot} 

\def\ie{\emph{i.e}\onedot}

\makeatother

\definecolor{gblue}{HTML}{4285F4}
\definecolor{gred}{HTML}{DB4437}
\definecolor{ggreen}{HTML}{0F9D58}
\definecolor{vblue}{HTML}{2993ba}

\definecolor{gbest}{HTML}{FFCCCB}
\definecolor{gsecond}{HTML}{FFE5CC}
\definecolor{gthird}{HTML}{FFF2A0}

\definecolor{todo}{RGB}{255,0,0}
\definecolor{revision}{RGB}{0,0,255}

\definecolor{colorMono}{rgb}{1.0, 0.9, 0.9}
\definecolor{colorMetric}{rgb}{1.0, 1.0, 0.8}
\definecolor{colorPrim}{rgb}{0.9, 0.9, 1.0}

\acrodef{sdf}[SDF]{signed distance function}
\acrodef{mlp}[MLP]{multi-layer perceptron}
\acrodef{nerf}[NeRF]{Neural Radiance Field}
\acrodef{3dgs}[3DGS]{3D Gaussain Splatting}
\acrodef{gt}[GT]{Ground Truth}
\acrodef{cd}[CD]{Chamfer Distance}
\acrodef{fscore}[F-score]{F-score}
\acrodef{nc}[NC]{Normal Consistency}
\acrodef{miou}[mIoU]{Mean Intersection over Union}
\acrodef{fr}[FR]{full-reference}
\acrodef{nr}[NR]{no-reference}
\acrodef{sds}[SDS]{Score Distillation Sampling}
\acrodef{sfm}[SfM]{Structure-from-Motion}
\acrodef{vfx}[VFX]{Visual effects}

\newcommand{\sota}{\text{state-of-the-art}\xspace}

\newcommand{\modelname}{\textbf{\textsc{DecomVoxel}}\xspace}
\newcommand{\trellis}{TRELLIS\xspace}
\newcommand{\supp}{\textit{supplementary}\xspace}
\newcommand{\fscore}{\mbox{F-Score}\xspace}

\usepackage{listings}

\begin{document}

\setcopyright{cc}
\setcctype{by}
\acmJournal{TOG}
\acmYear{2026}
\acmVolume{45}
\acmNumber{6}
\acmArticle{214}
\acmMonth{12}
\acmDOI{10.1145/3842522}

\title{\modelname: Harnessing 3D-Native Priors with Guided In-situ Denoising Optimization for Decompositional Scene Reconstruction}

\author{Junfeng Ni}
\authornote{Equal contribution}
\affiliation{%
  \institution{Tsinghua University, State Key Laboratory of General Artificial Intelligence, BIGAI}
  \city{Beijing}
  \country{China}
}
\email{njf23@mails.tsinghua.edu.cn}

\author{Zirui Zhou}
\authornotemark[1]
\affiliation{%
  \institution{Tsinghua University, State Key Laboratory of General Artificial Intelligence, BIGAI}
  \city{Beijing}
  \country{China}
}
\email{2033616887@qq.com}

\author{Yixin Chen}
\authornote{Corresponding authors}
\affiliation{%
  \institution{State Key Laboratory of General Artificial Intelligence, BIGAI}
  \city{Beijing}
  \country{China}
}
\email{ethanchen@g.ucla.edu}

\author{Yu Liu}
\affiliation{%
  \institution{Tsinghua University, State Key Laboratory of General Artificial Intelligence, BIGAI}
  \city{Beijing}
  \country{China}
}
\email{yu-liu24@mails.tsinghua.edu.cn}

\author{Nan Jiang}
\affiliation{%
  \institution{Peking University}
  \city{Beijing}
  \country{China}
}
\email{nan.jiang@stu.pku.edu.cn}

\author{Zhifei Yang}
\affiliation{%
  \institution{Peking University}
  \city{Beijing}
  \country{China}
}
\email{zhifei.yeung@gmail.com}

\author{Song-Chun Zhu}
\affiliation{%
  \institution{Tsinghua University, State Key Laboratory of General Artificial Intelligence, BIGAI}
  \city{Beijing}
  \country{China}
}
\email{s.c.zhu@pku.edu.cn}

\author{Siyuan Huang}
\authornotemark[2]
\affiliation{%
  \institution{State Key Laboratory of General Artificial Intelligence, BIGAI}
  \city{Beijing}
  \country{China}
}
\email{huangsiyuan@ucla.edu}

\begin{abstract}
  Decompositional scene reconstruction aims to reconstruct high-quality objects and background, yet existing methods still struggle with the level of quality under heavy occlusions. While generative priors offer a potential solution, 2D image-based priors often suffer from multi-view inconsistency due to a lack of 3D awareness. Conversely, 3D-native priors provide stronger structural inductive biases but frequently lead to spatial drift and misalignment within complex scenes. 
  To address these issues, we propose \modelname, formulating object completion as a guided in-situ denoising optimization that bridges 3D-native priors with neural scene reconstruction. Our framework introduces a reformulated epsilon-based distillation loss to ensure stable latent refinement, alongside adaptive spatial guidance that utilizes occupied and vacant anchors with temporal annealing to suppress generative hallucinations and mitigate spatial drift. Experiments on Replica and ScanNet++ show that \modelname significantly outperforms state-of-the-art methods while faithfully preserving the original spatial layout, structural fidelity, and style-consistent texture. Our method pushes the boundary of decompositional reconstruction by delivering high-quality textured meshes with clean topology, geometry, and appearance, providing a robust solution for the decompositional reconstruction of complex real-world scenes.
  Code is available at \url{https://github.com/DecomVoxel/DecomVoxel}.
\end{abstract}

\begin{CCSXML}
<ccs2012>
   <concept>
       <concept_id>10010147.10010178</concept_id>
       <concept_desc>Computing methodologies~Artificial intelligence</concept_desc>
       <concept_significance>500</concept_significance>
       </concept>
 </ccs2012>
\end{CCSXML}

\ccsdesc[500]{Computing methodologies~Artificial intelligence}

\keywords{Decompositional Scene Reconstruction, Multi-View Scene Reconstruction, 3D-Native Priors}

\begin{teaserfigure}
    \centering
    \includegraphics[width=\linewidth]{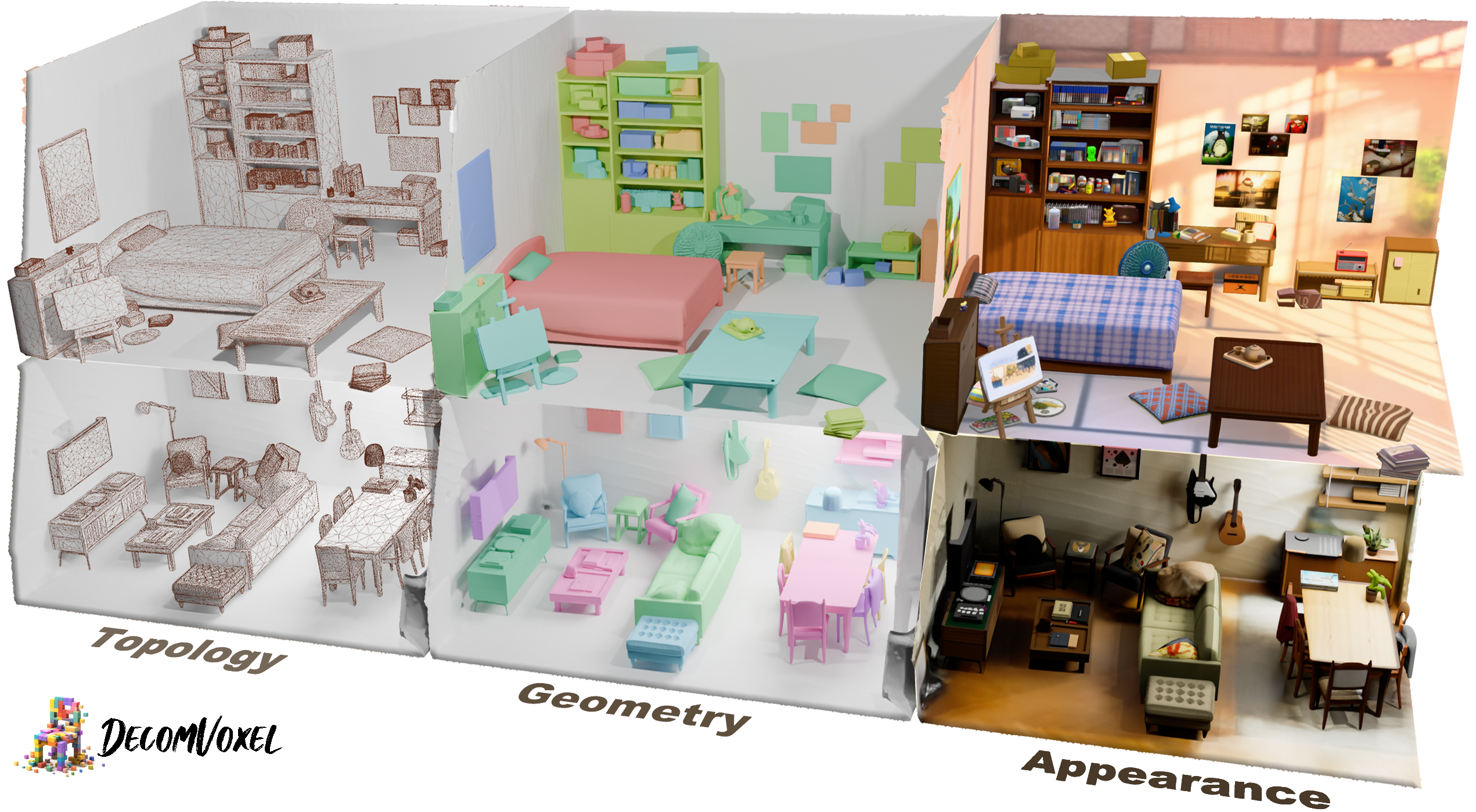}
    \caption{We propose \modelname, a framework that integrates guided 3D-native priors to enhance decompositional scene reconstruction. By formulating object completion as guided in-situ optimization, our method achieves high-quality topology, geometry, and appearance for both individual objects and backgrounds while faithfully preserving the original spatial layout.}
    \label{fig:teaser}
\end{teaserfigure}

\maketitle

% --------------------------- Section Separation Line --------------------------- 
\section{Introduction}

While rapid progress in neural 3D representations~\cite{mildenhall2020nerf,kerbl3Dgaussians} has revolutionized holistic scene reconstruction, recent approaches~\cite{wu2022object,wu2023objsdfplus,kong2023vmap,li2023rico} have shifted focus towards \textit{decompositional reconstruction} to recover individual object entities, facilitating fine-grained interaction for downstream embodied AI tasks~\cite{li2024ag2manip,lu2025gwm}.
However, these methods struggle with incomplete geometry and incoherent appearances, due to inevitable occlusions and limited sensor viewpoints in real-world environments. 

To address these challenges, recent methods~\cite{ni2025decompositional,yang2025instascene} leverage generative priors from pre-trained image diffusion models~\cite{rombach2022high,liu2023zero} to complete missing visual and geometric content. However, due to their view-independent formulation and lack of 3D awareness, these \textit{image-based 2D priors} often fail to produce coherent 3D objects or consistent appearances. In contrast, another line of work~\cite{yu2025metascenes,xia2026simrecon,chen2025sam} employs 3D generative models~\cite{zhang20233dshape2vecset,lai2025hunyuan3d,li2025triposg} that encode geometric and structural priors from large-scale 3D datasets~\cite{deitke2023objaverse} for object completion. 
Nevertheless, despite better capturing coherent 3D geometry, these \textit{3D-native priors} struggle to maintain precise spatial alignment and texture fidelity within complex scene contexts, due to the lack of explicit guidance and global constraints to regularize the inherent stochasticity of generative priors.

In this paper, we propose \modelname, a framework that bridges this gap by formulating the completion of missing regions in decompositional reconstruction as a \textit{guided in-situ denoising optimization} to harness 3D-native priors with neural scene reconstruction. 
We design the optimization pipeline that leverages structured scene representations, \ie, sparse voxel~\cite{sun2025sparse}, which provides flexibility to seamlessly connect with state-of-the-art 3D object generation models~\cite{xiang2025structured}. To effectively leverage the generative power, we propose a reformulated epsilon-based distillation loss for flow matching, where we demonstrate that optimizing the noise residual implicitly introduces a principled temporal weighting to the optimizing gradient of the latent feature. This formulation improves convergence stability and detailed geometry recovery compared with standard velocity-based optimization.

To regularize the inherent stochasticity of generative priors, which can lead to spatial drift and geometric misalignment, we introduce \textit{adaptive spatial guidance} to steer the optimization through non-uniform spatial constraints and temporal annealing. Specifically, we define occupied anchors within regions of high-confidence reconstruction to preserve coherence with observations, and vacant anchors where a re-distributed pruning strategy prevents the generative model from hallucinating in zones where the object must not exist. We employ an annealing schedule that dynamically decays noise levels and guidance weights to balance generative capacity and spatial constraints, facilitating structural completion in early stages and fine-grained refinement as the latent state converges toward the clean manifold.

We evaluate our method on the Replica~\cite{replica19arxiv} and ScanNet++~\cite{yeshwanthliu2023scannetpp} datasets. Comprehensive experiments demonstrate that \modelname consistently outperforms \sota baselines in both geometry and appearance quality, particularly in challenging, heavily occluded scenarios. Unlike existing baselines that suffer from geometric distortion and scene-level spatial misalignment, our approach faithfully preserves the original spatial layout, structural fidelity, and style-consistent texture. Notably, our method produces high-fidelity textured meshes with \textit{clean topology, complete geometry, and consistent appearance}, providing a robust solution for the decompositional reconstruction of complex real-world scenes, as shown in \cref{fig:teaser}. 

In summary, our main contributions are as follows:
\begin{itemize}
    \item We propose \modelname, formulating decompositional reconstruction as an \textit{guided in-situ denoising optimization} that integrates 3D-native priors with neural scene reconstruction.
    \item We introduce an epsilon-based distillation objective coupled with adaptive spatial guidance, which applies spatial regularization through temporal annealing to effectively suppress generative hallucinations and mitigate spatial drift.
    \item Extensive experiments on Replica and ScanNet++ demonstrate that \modelname achieves state-of-the-art performance. Our framework consistently produces high-fidelity results with clean topology, geometry, and appearance, striking a principled balance between neural scene reconstruction and 3D-native priors.
\end{itemize}

\section{Related Work}

\subsection{3D Scene Representation and Reconstruction}

The rapid advancement of neural implicit representations, such as \ac{nerf}~\cite{mildenhall2020nerf} and \ac{3dgs}~\cite{kerbl3Dgaussians}, has significantly enhanced the fidelity of scene reconstruction from 2D images~\cite{muller2022instant,yu2022monosdf,yang2024gurecon,li2023neuralangelo,lu2024scaffold,huang20242d,chen2024pgsr}. To further provide more structured and manageable 3D representations, recent works have explored sparse voxel-based differentiable rendering~\cite{sun2025sparse,li2025geosvr,oh2025svrecon,chi2026advancing}. 
To facilitate downstream interactions such as object editing~\cite{chen2024gaussianeditor,li2025voxhammer}, manipulation~\cite{lu2024manigaussian,li2024ag2manip}, and simulation~\cite{xie2024physgaussian,li2023pac}, substantial efforts~\cite{wu2023objsdfplus,li2023rico,lyu2024total,ye2024gaussian,gu2024egolifter,shen2025trace3d} have shifted towards decompositional scene reconstruction, which aims to extract individual 3D representations for each object and the background.
However, existing methods typically struggle to recover complete geometries, particularly in occluded regions. To address this limitation, we integrate a 3D-native prior into a voxel-based neural reconstruction framework, enabling the model to recover missing structures in situ while faithfully preserving scale and spatial alignment.

\subsection{Generative Priors for Decompositional Reconstruction}
To achieve complete object reconstruction in occluded areas, several methods~\cite{chung2023luciddreamer,huang2025midi,yao2025cast,dong2025hiscene} have explored incorporating generative priors into decompositional reconstruction. Some works~\cite{chen2023fantasia3d,liu2023zero,qiu2024richdreamer,weber2024nerfiller} utilize 2D diffusion models to fill missing 3D information; however, they often struggle with multi-view inconsistency~\cite{ni2025decompositional,yang2025instascene,xia2025holoscene}, leading to the Janus problem~\cite{shi2023mvdream,lin2023magic3d}. With the recent flourish of 3D-native generative models~\cite{hong2023lrm,zhang20233dshape2vecset,ye2025hi3dgen,xiang2025native}, another branch of research~\cite{meng2025scenegen,chen2026scorp,dong2026replicateanyscene,lin2026mix3r,zadaianchuk2026reconstruction} seeks to ensure inherent 3D consistency by directly leveraging these 3D-native priors. For instance, MetaScenes~\cite{yu2025metascenes} and SimRecon~\cite{xia2026simrecon} utilize image-to-3D models~\cite{tochilkin2024triposr,zhang2024clay,lai2025hunyuan3d} to generate meshes, while Amodal3R~\cite{wu2025amodal3r} incorporates masks as auxiliary conditions to enhance generation quality under heavy occlusion.
However, their resulting geometry and textures frequently deviate from the original observations by using 3D generative models as separate modules. Furthermore, these methods require a post-hoc alignment process to transform the meshes back into the scene coordinates. Together, these limitations lead to cumulative errors that degrade overall reconstruction quality. To address them, our method reformulates the completion of missing regions as an in-situ denoising optimization rather than a pipeline of decoupled generation and alignment. By optimizing directly within the scene coordinates, we eliminate the need for error-prone alignment and ensure seamless spatial consistency between the generated and observed regions.

\subsection{Guidance for Denoising Optimization}
Extensive research~\cite{wang2023prolificdreamer,zhang2023adding,ye2023ip,lu2025movis,zhong2025taming} has explored incorporating guidance during the denoising process to enhance fidelity relative to original inputs. Recent efforts~\cite{yang2025holopart,zhou2025amodalgen3d,li2026worldgrow,huang2026unirecgen,li2026mv} have transitioned to the 3D domain. For instance, ReconViaGen~\cite{chang2025reconviagen} Geometry Forcing~\cite{wu2025geometry} and CUPID~\cite{huang2025cupid} inject implicit features derived from pre-trained multi-view stereo models~\cite{wang2025vggt} into generative pipelines to improve 3D consistency, but retain alignment errors due to spatial ambiguity and inaccurate predictions. In contrast, approaches like SAM3D~\cite{chen2025sam} and ShapeR~\cite{siddiqui2026shaper} introduce explicit object point clouds as conditional inputs to constrain the spatial content. However, these feed-forward methods typically lack the per-scene precision required for complex environments. Unlike these approaches, our method employs an in-situ denoising optimization strategy driven by explicit spatial guidance. By treating explicit 3D guidance as persistent supervisory signals within an iterative loop, our framework progressively recovers missing textures and structures while adhering to the spatial evidence of the 3D scenes.

\begin{figure*}[t]
  \includegraphics[width=\linewidth]{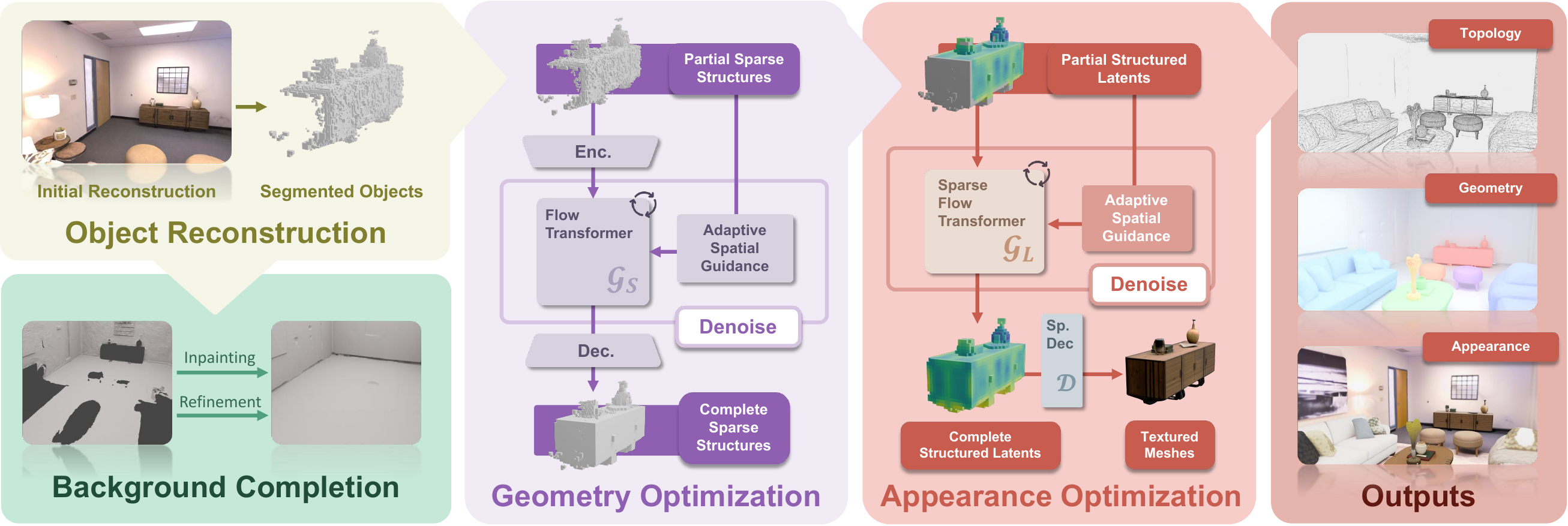}
  \caption{\textbf{Overview of \modelname.} Our framework bridges 3D generative priors with neural reconstruction through a guided in-situ denoising optimization. We segment objects from the initial scene and then sequentially optimize their geometry and appearance under adaptive spatial guidance. This process recovers missing structures and consistent textures while preserving the original scene layout, producing high-fidelity topology, geometry, and appearance.}
  \label{fig:overview}
\end{figure*}

% --------------------------- Section Separation Line --------------------------- 
\section{Method}
In this section, we first establish the basics of 3D representations and generative priors in \cref{sec:preliminaries}. \cref{sec:bridge_prior} introduces a reformulated in-situ denoising optimization that bridges initial scene segments with 3D-native generative priors. To effectively regularize the optimization, \cref{sec:guidance} describes an adaptive guidance strategy using spatial anchors and temporal annealing. \cref{sec:optimization_pipeline} details the overall pipeline of holistic object and background reconstruction.

\subsection{Preliminaries}
\label{sec:preliminaries}

Our framework is built upon two design choices: a voxel-based scene representation and a 3D-native generative prior. 
The use of a sparse voxel representation for decompositional reconstruction ensures a seamless connection with 3D generative priors, which operate on a structured voxel latent space. Furthermore, voxels provide a natural basis for incorporating our explicit 3D guidance introduced in~\cref{sec:guidance}, allowing for precise and direct geometric anchoring throughout the optimization process.
Specifically, we adopt \trellis~\cite{xiang2025structured} as our generative prior, as its 3D-native architecture inherently ensures multi-view consistency. \trellis also provides joint modeling of both geometry and appearance, enabling our framework to simultaneously complete missing structures and synthesize consistent textures. 

\paragraph{Sparse Voxel Representation} Following SVRaster~\cite{sun2025sparse} and GeoSVR~\cite{li2025geosvr}, we represent the 3D scene using a set of explicit sparse voxels $\mathcal{V} = \{v_i\}_{i=1}^N$ organized in an octree structure. Each voxel $v_i$ is characterized by its center $c_i$, size $s_i$, density $\sigma_i$, and spherical harmonic coefficients $sh_i$ for appearance. 
For each object $k$, we identify its corresponding sparse voxel set $\mathcal{V}_k \subset \mathcal{V}$ by fusing the 2D instance masks through volume fusion. Specifically, a voxel $v_i$ is assigned to $\mathcal{V}_k$ if its projected 2D location consistently falls within the instance mask of object $k$ across multiple views.
Furthermore, we utilize the geometric uncertainty $U_{geom}$ introduced in GeoSVR to quantify the confidence of the initial reconstruction, thereby identifying deterministic regions that serve as reliable spatial anchors.

\paragraph{3D-Native Prior}
Our framework specifically utilizes three specialized modules from \trellis~\cite{xiang2025structured}: a sparse structure generation module $G_S$, a latent refinement module $G_L$, and a pre-trained VAE decoder $\mathcal{D}_{vae}$.
The structure module $G_S$ comprises an encoder $\mathcal{E}$, a decoder $\mathcal{D}$, and a structural flow model $\mathbf{v}_S$. The encoder $\mathcal{E}$ maps initial 3D representations (\eg, incomplete voxel sets $\mathcal{V}_k$) into the latent space. $\mathbf{v}_S$ iteratively refines the coarse geometric latents, and the decoder $\mathcal{D}$ reconstructs the final geometry from the refined latents. The module $G_L$ primarily governs the synthesis of object appearance by employing a latent flow model $\mathbf{v}_L$ to denoise the structured latents. Finally, the refined structured latents are passed through the VAE decoder $\mathcal{D}_{vae}$ to produce the final high-quality object mesh and its corresponding appearance.
For both $G_S$ and $G_L$, the generation process is formulated as a conditional flow matching (CFM) problem~\cite{lipman2022flow}. Given a data distribution $x_0$ and noise $\epsilon$, the model learns a vector field $v_\theta(x_t, t)$ to approximate the target velocity along the probability path $x_t = (1-t)x_0 + t\epsilon$:
\begin{equation}
\label{eq:cfm}
    \mathcal{L}_{\theta} = \mathbb{E}_{t, x_0, \epsilon} \| v_\theta(x_t, t) - (\epsilon - x_0) \|_2^2.
\end{equation}

\begin{figure}[t]
  \includegraphics[width=\linewidth]{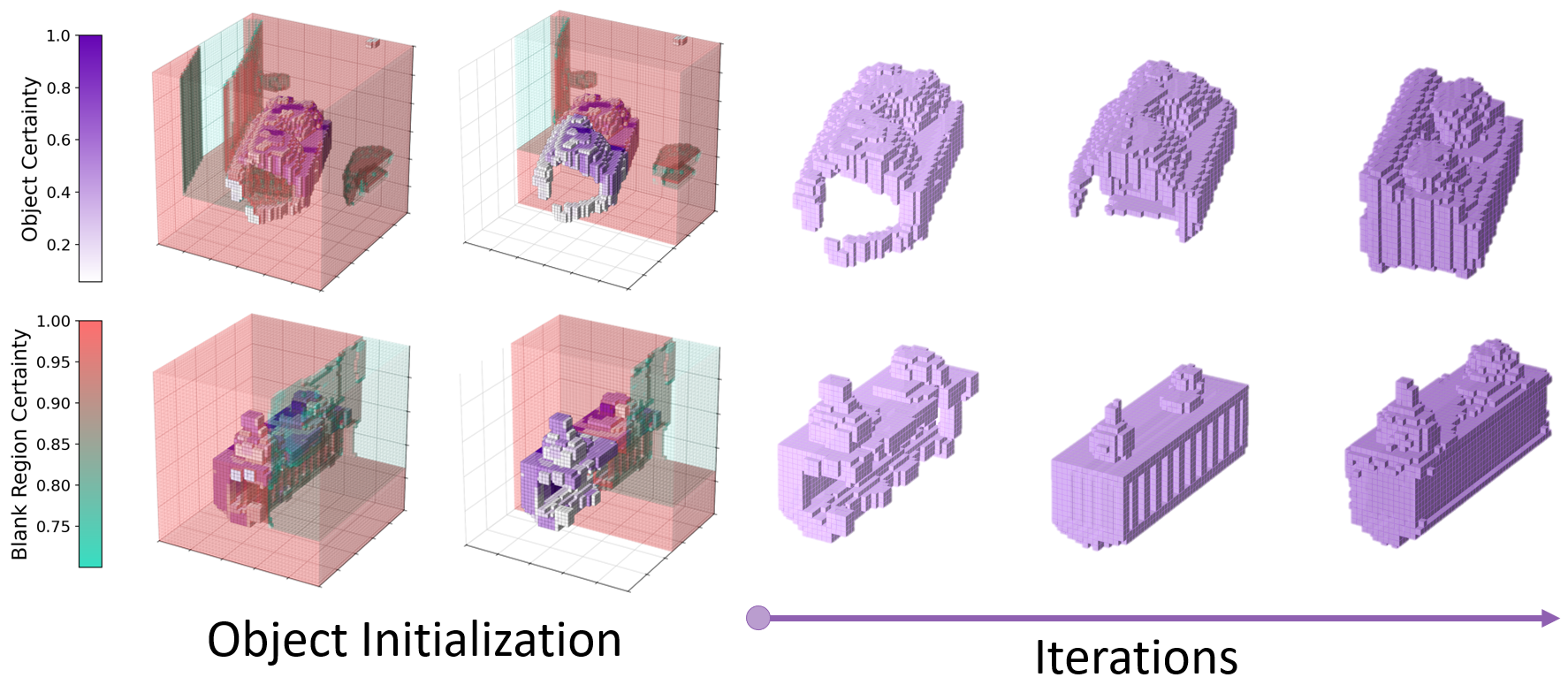}
  \caption{\textbf{Visualization of intermediate results.} The left panel illustrates our spatial anchors: the occupied anchor $\Omega_O$ (purple) preserves verified structures, while the vacant anchor $\Omega_V$ (red) suppresses hallucinations in prohibited zones. The green regions denote the free space for structural synthesis. The right panel displays the progressive geometric completion over iterations. Guided by these deterministic anchors, the generative process restores missing regions while maintaining the original scene layout.}
  \label{fig:intermediate_results}
\end{figure}

\subsection{In-situ Denoising Optimization}
\label{sec:bridge_prior}

A primary objective in decompositional reconstruction is to recover the complete geometry and appearance of individual objects from incomplete initial segments $\{\mathcal{V}_k\}$. We formulate this completion as an \textit{in-situ denoising optimization} within the scene-level coordinate space, leveraging a 3D-native generative prior.

Given that initial geometries are typically noisy and incomplete, a straightforward approach is to encode the initial segment $\mathcal{V}$ into a latent embedding $x'$ by the encoder $\mathcal{E}$ and leverage the pre-trained Flow Matching model $v_\theta$ to iteratively update $x'$ via gradient-based denoising. Analogous to score-based refinement in diffusion models, this process uses the velocity field to minimize the discrepancy between the predicted velocity and the target vector field: 
\begin{equation}
\label{eq:v_loss}
    \mathcal{L}_{v} = \mathbb{E}_{t, \epsilon} \left[\| v_\theta(x_t, t) - (\epsilon - x') \|_2^2 \right],
\end{equation}
where $x_t = (1-t)x' + t\epsilon$ represents the probability path at timestep $t$. However, we empirically observe that directly optimizing \cref{eq:v_loss} often leads to unstable convergence or over-smoothed results. This stems from the fact that $\mathcal{L}_{v}$ treats all noise levels with uniform gradient importance, failing to account for the varying structural reliability along the denoising trajectory.

To address this, we rethink the distillation process by drawing inspiration from the \ac{sds}~\cite{poole2022dreamfusion} paradigm. We propose a reformulated epsilon-based distillation loss for Flow Matching:
\begin{equation}
\label{eq:epsilon_loss}
    \mathcal{L}_{\epsilon} = \mathbb{E}_{t, \epsilon} \left[\| \hat{\epsilon}_\theta(x_t; t) - \epsilon \|_2^2 \right],
\end{equation}
where $\hat{\epsilon}_\theta(x_t; t) = x_t + (1-t)v_\theta(x_t, t)$ is the noise residual estimated from the velocity field. By substituting the definition of $x_t$, we can express the residual as $\hat{\epsilon}_\theta = (1-t)x' + t\epsilon + (1-t)v_\theta$. 

We formally prove that our proposed objective $\mathcal{L}_{\epsilon}$ and the velocity loss $\mathcal{L}_{v}$ share identical gradient directions with respect to the latent $x'$, differing only by a time-dependent scaling factor: $\nabla_{x'} \mathcal{L}_{\epsilon} = (1-t) \nabla_{x'} \mathcal{L}_{v}$ (see \supp for the full derivation). This $(1-t)$ factor arises naturally from $\frac{\partial x_t}{\partial x'} = (1-t)$, which serves as a principled adaptive scheduler. At high noise levels ($t \to 1$), where the generative power is more stochastic, the gradient magnitude is naturally suppressed to reduce erroneous geometric deformations. Conversely, as the latents converge toward a cleaner manifold ($t \to 0$) where the model's velocity predictions are most accurate, the gradient magnitude automatically increases, providing stronger, more precise guidance for fine-grained refinement.

\begin{figure*}[t]
  \includegraphics[width=\linewidth]{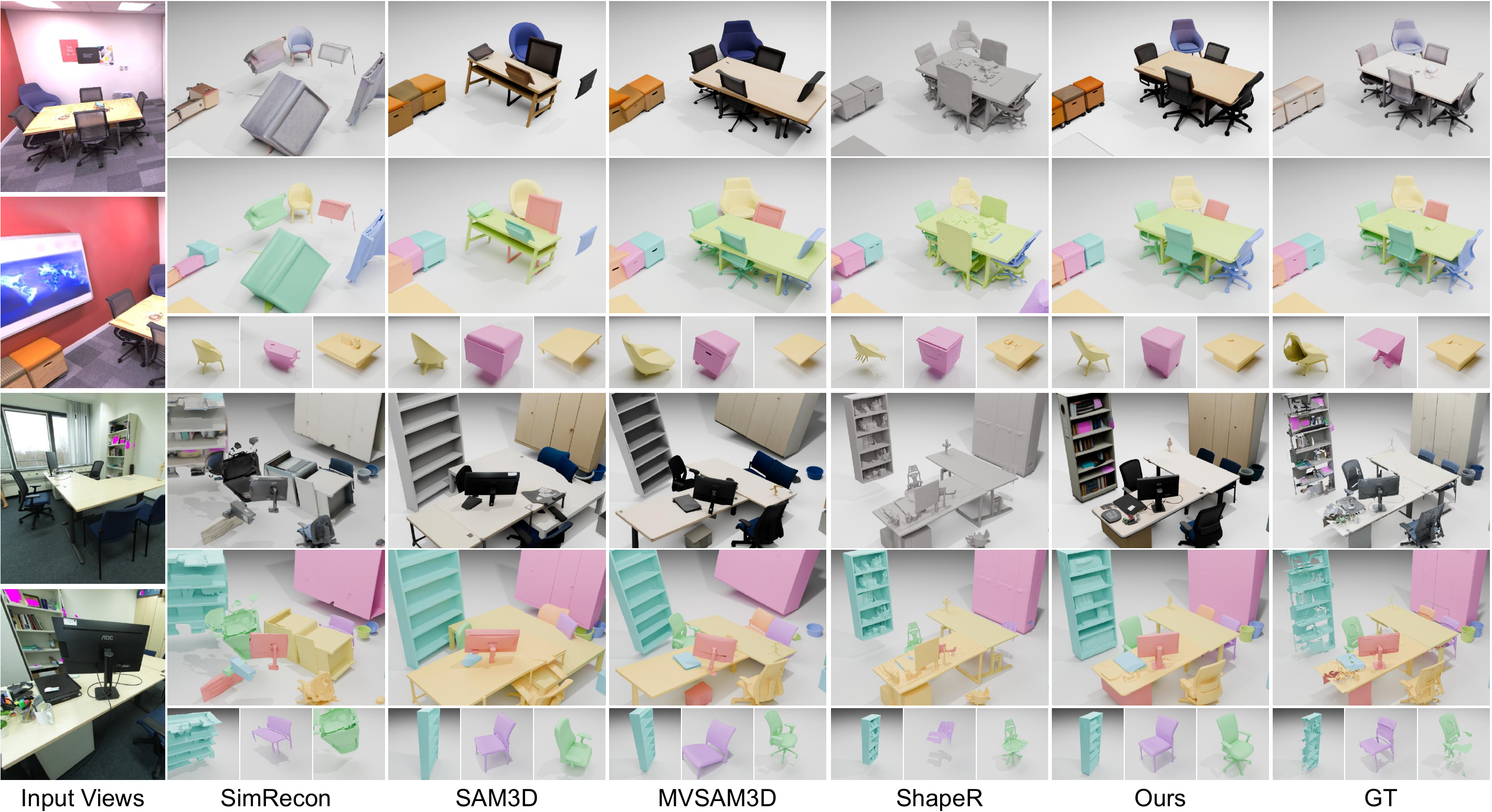}
  \caption{\textbf{Qualitative comparison.} Our method reconstructs complete geometry and consistent appearance while faithfully preserving the original scene layout, whereas baselines exhibit substantial drift in position, rotation, and scale.}
  \label{fig:qual_results}
\end{figure*}

\subsection{Adaptive Spatial Guidance}
\label{sec:guidance}

To regularize the inherent stochasticity of 3D-native generative models, which often compromise the spatial fidelity of reconstructed objects, we propose an \textit{adaptive spatial guidance} strategy to anchor the denoising process using spatial constraints and dynamically balance generative completion and structural preservation.

\paragraph{Spatial-Anchored Guidance}
We enforce spatial fidelity by identifying two types of anchoring zones for each object: the \textit{occupied anchor} ($\Omega_O$), representing regions where geometry is confidently reconstructed, and the \textit{vacant anchor} ($\Omega_V$), representing regions where the object must not exist. Critically, the latent representation $x'$ in \trellis resides in a structured voxel feature space, where each voxel is associated with a 3D location through a deterministic mapping from real-world Euclidean coordinates to the latent feature. This spatial correspondence allows us to apply guidance directly to the latent features based on the underlying scene geometry.

For the occupied anchor $\Omega_O$, we filter voxels from the initial reconstruction $\mathcal{V}$ that exhibit low geometric uncertainty in $U_{geom}$ (defined in \cref{sec:preliminaries}).
We thus propose a structural preservation loss $\mathcal{L}_{O}$ that regularizes the optimized latent features to remain consistent with the initial features:
\begin{equation}
    \mathcal{L}_O(x') = \sum_{\mathbf{v} \in \Omega_O} w_O \| x'(\mathbf{v}) - x'_{init}(\mathbf{v}) \|_2^2,
\end{equation}
where  $x'(\mathbf{v})$ denotes the latent feature at voxel position $\mathbf{v}$ and $x'_{init}$ is the corresponding initial value.
This loss is applied during both geometry and appearance optimization to stabilize the structure and observed textures, respectively. 

While $\Omega_O$ preserves existing structures, it cannot prevent the generative prior from hallucinating unwanted geometry in empty spaces. We therefore define the vacant anchor $\Omega_V$ by fusing two sources of information: 1) free-space regions identified via ray-casting from input depth maps, and 2) regions occupied by the background or other objects. 
A naive approach would be to prune the features in $\Omega_V$ to zero. However, we observe that within the sparse latent space of the pre-trained prior, "inactive" regions—those representing empty space—do not possess zero-valued features. Instead, they follow a specific stochastic distribution characterized by a low-magnitude mean. Forcing these features to zero would push the latent state $x'$ off the learned manifold, leading to optimization instability and rendering artifacts.

To address this, we propose \textit{a re-distributed pruning strategy} that nudges features in $\Omega_V$ toward the "void prior." For each voxel $\mathbf{v} \in \Omega_V$, we compute a target mean $\mu_{new}$ by interpolating the current feature mean $\mu$ toward a predefined near-zero prior mean $\mu_0$:
\begin{equation}
\label{eq:mean_redistribution}
\mu_{new} = (1 - \lambda_p) \mu + \lambda_p \mu_0,
\end{equation}
where $\lambda_p \in [0, 1]$ is a redistribution factor. To maintain the stochasticity expected by the VAE decoder, the latent features $x'(\mathbf{v})$ in $\Omega_V$ are then re-sampled from a uniform distribution:
\begin{equation}
\label{eq:resampling}
x'(\mathbf{v}) \sim \mathcal{U}(\mu_{new} - r, \mu_{new} + r), \quad \forall \mathbf{v} \in \Omega_V,
\end{equation}
where $r$ represents a small distributional radius. This strategy suppresses extra structure growth in prohibited zones while ensuring the latent features remain compatible within the 3D prior manifold, resulting in cleaner boundaries and more stable convergence.

\begin{algorithm}[t]
\caption{Overall optimization pipeline of \modelname.}
\label{alg:decomvoxel}
\DontPrintSemicolon
\KwIn{Posed RGB images $\{I_j,P_j\}_{j=1}^{J}$ and instance masks $\{M_j\}_{j=1}^{J}$}
\KwOut{Object meshes $\{\mathcal{M}_k\}_{k=1}^{K}$ and background mesh $\mathcal{B}$}

\tcp{Stage I: Decompositional Reconstruction}
$\mathcal{V}\leftarrow\operatorname{GeoSVR}(\{I_j,P_j\}_{j=1}^{J})$\;
$\{\mathcal{V}_k\}_{k=1}^{K},\mathcal{V}_{\mathrm{bg}}\leftarrow
\operatorname{Partition}(\mathcal{V},\{M_j,P_j\}_{j=1}^{J})$\;
Estimate the geometric uncertainty $U_{\mathrm{geom}}$ for the reconstructed voxels $\mathcal{V}$\;

\tcp{Stage II: Background Completion}
$\{I_j^{\mathrm{bg}}\}\leftarrow\operatorname{See3D}(\{I_j\},\{M_j\})$\;
Fit dominant 3D planes and render depth supervision $\{D_j^{\mathrm{bg}}\}$\;
Re-optimize $\mathcal{V}_{\mathrm{bg}}$ using $\{I_j^{\mathrm{bg}},D_j^{\mathrm{bg}},P_j\}$\;
$\mathcal{B}\leftarrow\operatorname{MarchingCubes}(\mathcal{V}_{\mathrm{bg}})$\;

\tcp{Stage III: In-situ Object Refinement}
\ForEach{$\mathcal{V}_k$, $k=1,\ldots,K$}{
    Construct occupied and vacant anchors $\Omega_O^k$ and $\Omega_V^k$\;
    Initialize the learnable occupancy logits $\mathbf{Z}_\theta$ from $\mathcal{V}_k$\;
    Define $\mathbf{S}_\theta\equiv\operatorname{sigmoid}(\mathbf{Z}_\theta)$\;

    \tcp{Geometry Optimization}
    \For{$q\leftarrow1$ \KwTo $N_G$}{
        $x'_S\leftarrow\mathcal{E}(\mathbf{S}_\theta)$\;
        Sample $t$ from the scheduled timestep range and $\epsilon\sim\mathcal{N}(0,I)$\;
        $x_{S,t}\leftarrow(1-t)x'_S+t\epsilon$\;
        $\hat{\epsilon}_S\leftarrow x_{S,t}+(1-t)\mathbf{v}_S(x_{S,t},t)$\;
        Update $\theta$ using $\mathcal{L}_{\epsilon}^{\mathrm{geo}}+w_O(q)\mathcal{L}_{O}^{\mathrm{geo}}$\;
        \If{$q\bmod 300=0$ and $q\le N_V$}{
            Redistribute $x'_S$ within $\Omega_V^k$ using $\lambda_p(q)$\;
        }
    }
    $\mathcal{V}_k^{\mathrm{com}}\leftarrow
    \operatorname{Threshold}\!\left(\mathcal{D}(\mathcal{E}(\mathbf{S}_\theta)),\tau\right)$\;

    \tcp{Appearance Optimization}
    Initialize $x'_L$ by aggregating multi-view DINOv2 features on $\mathcal{V}_k^{\mathrm{com}}$\;
    \For{$q\leftarrow1$ \KwTo $N_A$}{
        Sample $t$ from the scheduled timestep range and $\epsilon\sim\mathcal{N}(0,I)$\;
        $x_{L,t}\leftarrow(1-t)x'_L+t\epsilon$\;
        $\hat{\epsilon}_L\leftarrow x_{L,t}+(1-t)\mathbf{v}_L(x_{L,t},t)$\;
        Update $x'_L$ using $\mathcal{L}_{\epsilon}^{\mathrm{app}}+w_A(q)\mathcal{L}_{O}^{\mathrm{app}}$\;
    }
    $\mathcal{M}_k\leftarrow\mathcal{D}_{\mathrm{vae}}(x'_L)$\;
}
\Return{$\{\mathcal{M}_k\}_{k=1}^{K}$ and $\mathcal{B}$}\;
\end{algorithm}

\paragraph{Temporal Annealing Strategy}
An inherent tension exists between the 3D-native prior and spatial guidance: high generative power is required to synthesize complex missing regions, yet it risks distorting the existing spatial structure. Conversely, aggressive guidance ensures spatial agreement but may suppress the synthesis of novel geometry. To resolve this, we employ a temporal annealing strategy that dynamically modulates the noise timestep $t$ and the guidance weight $w_O$ for $\mathcal{L}_{O}$ throughout the optimization process.

In the early iterations, we sample timesteps $t$ from higher ranges to encourage the generative power. Simultaneously, the spatial guidance weights $w_O$ are maintained at their maximum values to anchor the generative process, ensuring the existing high-confidence segments $\Omega_O$ remain strictly anchored and the vacant zones $\Omega_V$ remain suppressed. As the optimization progresses, we implement an annealing schedule: the lower bound of the sampled $t$ is linearly decayed to shift the focus toward local geometric refinement. Concurrently, $w_O$ and $\lambda_p$ follow a cosine decay curve to relax the spatial constraints. In the final iterations, the structural preservation loss $\mathcal{L}_O$ and the pruning strategy for $\Omega_V$ are deactivated ($w_O \to 0$).
By removing the rigid anchors, the prior can seamlessly integrate the newly synthesized structures with the observed features, ensuring manifold consistency and global coherence without the artifacts associated with over-constrained optimization. More details are provided in \supp.

\subsection{Optimization Pipeline}
\label{sec:optimization_pipeline}

Building upon the generative priors and guidance mechanisms defined above, our optimization pipeline proceeds in sequential stages. 
\Cref{alg:decomvoxel} summarizes the overall framework.

\paragraph{Stage I: Decompositional Reconstruction}
We first perform an initial scene reconstruction using GeoSVR~\cite{li2025geosvr} to obtain the global sparse voxel set $\mathcal{V}$, and then partition the scene into individual object voxel sets $\{\mathcal{V}_k\}$ and background $\mathcal{V}_{bg}$ with instance masks. This stage establishes the initial spatial anchors and geometric uncertainty $U_{geom}$ for subsequent optimization.

\paragraph{Stage II: Background Completion}
Since 3D-native generative models are primarily trained on object-level datasets,
we use See3D~\cite{ma2025see3d}, a video-based diffusion model, to inpaint color in unobserved background regions. 
Following G4Splat~\cite{ni2026g4splat}, we additionally fit 3D planes to the background to provide consistent depth supervision for both observed and unobserved areas. 
The background voxel representation is then re-optimized using the inpainted color and depth maps, after which Marching Cubes~\cite{lorensen1998marching} is applied to extract a complete background mesh.

\paragraph{Stage III: In-situ Object Refinement}
For each initial object $\mathcal{V}_k$, we perform two-stage optimization to recover complete geometry and high-fidelity appearance, conditioned on a reference image in which occluded regions are completed and the object is recentered by a multimodal image refinement model~\cite{seedream2025seedream}.
We first recover the complete geometric structure by optimizing a learnable continuous structural grid $\mathbf{S}_\theta \in [0, 1]$, which is initialized from $\mathcal{V}_k$. This grid is mapped by the encoder $\mathcal{E}$ to the latent representation $x'_S = \mathcal{E}(\mathbf{S}_\theta)$. We optimize the structural parameters $\theta$ by minimizing:
\begin{equation}
\label{eq:total_geo_loss}
\mathcal{L}_{total}^{geo} = \mathcal{L}_{\epsilon}^{geo}(x'_S) + \mathcal{L}_{O}(x'_S),
\end{equation}
where $\mathcal{L}_{\epsilon}^{geo}$ denotes the epsilon-based distillation loss guided by the structural module $G_S$, and $\mathcal{L}_{O}$ denotes the structural preservation loss. During this stage, we apply the re-distributed pruning strategy within $\Omega_V$ to suppress hallucinated geometry in prohibited regions. Upon convergence, the decoder $\mathcal{D}$ maps the refined latents to the completed voxel geometry $\mathcal{V}_k^{com}$.

We then optimize voxelized appearance latents $x'_L$ to synthesize high-fidelity textures for the unobserved regions of $\mathcal{V}_k^{com}$. These appearance latents are initialized by back-projecting DINOv2~\cite{oquab2023dinov2} features from observed multi-view images into the voxel space. The appearance is refined by minimizing
\begin{equation}
\label{eq:total_app_loss}
\mathcal{L}_{total}^{app} = \mathcal{L}_{\epsilon}^{app}(x'_L) + \mathcal{L}_{O}(x'_L).
\end{equation}
Here, the generative guidance is provided by the appearance flow model $G_L$. By maintaining the spatial anchor $\mathcal{L}_{O}$ during this stage, the synthesized textures remain consistent with the observed textures on the initial object segments. 
More details on spatial-guidance construction, scheduling, and optimization parameters are provided in \supp.

% --------------------------- Section Separation Line --------------------------- 
\section{Experiments}
We evaluate \modelname in terms of geometric reconstruction, appearance quality, and spatial fidelity. Additional details on the user-study protocol and significance tests, as well as limitations, are provided in \supp.

\begin{table}[t]
\centering
\small
\caption{\textbf{Quantitative comparison.} Our method consistently outperforms state-of-the-art baselines across both reconstruction and rendering metrics. Top-3 results are highlighted as the \colorbox{gbest}{first}, \colorbox{gsecond}{second} and \colorbox{gthird}{third}. }
\label{tab:main_results}
\resizebox{\linewidth}{!}{
\begin{tabular}{ll ccc ccc}
\toprule
\multirow{2}{*}{Dataset} & \multirow{2}{*}{Method} & \multicolumn{3}{c}{Reconstruction} & \multicolumn{3}{c}{Rendering} \\
\cmidrule(lr){3-5} \cmidrule(lr){6-8}
& & CD $\downarrow$ & F-score $\uparrow$ & NC $\uparrow$ & PSNR $\uparrow$ & SSIM $\uparrow$ & LPIPS $\downarrow$ \\
\midrule
\multirow{7}{*}{Replica} 
& SimRecon  & 17.55 & 40.61 & 63.25 & 18.04 & \colorbox{gthird}{0.858} & 0.152 \\
& ReconViaGen & 15.49 & \colorbox{gthird}{51.02} & \colorbox{gthird}{68.09} & 18.58 & \colorbox{gsecond}{0.861} & 0.147 \\
& CUPID & 16.29 & 43.95 & 64.80 & 16.70 & 0.809 & 0.180 \\
& SAM3D    & \colorbox{gthird}{10.59} & 45.46 & 66.33 & \colorbox{gsecond}{18.89} & 0.847 & \colorbox{gthird}{0.143} \\
& MV-SAM3D  & 12.87 & 40.46 & 64.19 & \colorbox{gthird}{18.83} & 0.848 & \colorbox{gsecond}{0.142} \\
& ShapeR  & \colorbox{gsecond}{6.59} & \colorbox{gsecond}{61.76} & \colorbox{gsecond}{69.09} & - & - & - \\
& Ours      & \colorbox{gbest}{3.95} & \colorbox{gbest}{77.87} & \colorbox{gbest}{78.85} & \colorbox{gbest}{24.21} & \colorbox{gbest}{0.907} & \colorbox{gbest}{0.114} \\
\midrule
\multirow{7}{*}{ScanNet++} 
& SimRecon  & 17.11 & 36.64 & 59.21 & 16.40 & 0.826 & 0.198 \\
& ReconViaGen & \colorbox{gthird}{11.75} & \colorbox{gthird}{45.36} & \colorbox{gsecond}{63.58} & \colorbox{gsecond}{17.59} & \colorbox{gsecond}{0.846} & 0.186 \\
& CUPID & 14.70 & 39.51 & \colorbox{gthird}{61.78} & 14.18 & 0.765 & 0.239 \\
& SAM3D    & 12.75 & 32.23 & 60.70 & 16.46 & 0.833 & \colorbox{gthird}{0.169} \\
& MV-SAM3D  & 15.35 & 26.95 & 60.51 & \colorbox{gthird}{17.41} & \colorbox{gthird}{0.842} & \colorbox{gsecond}{0.167} \\
& ShapeR  & \colorbox{gsecond}{9.77} & \colorbox{gsecond}{52.44} & 60.74 & - & - & - \\
& Ours      & \colorbox{gbest}{4.34} & \colorbox{gbest}{77.93} & \colorbox{gbest}{76.39} & \colorbox{gbest}{21.71} & \colorbox{gbest}{0.893} & \colorbox{gbest}{0.133} \\
\bottomrule
\end{tabular}
}
\end{table}

\begin{table}[t]
  \centering
  \caption{\textbf{User Study.} Based on 44 expert questionnaires, we assess the scores (rated 1–5) of object completeness and consistency in geometry and appearance, along with spatial-layout fidelity relative to the original scene. Our method significantly outperforms all baselines across all metrics, particularly in preserving global structural alignment.}
  \label{tab:user_study}
  \small
    \begin{tabular}{lccc}
    \toprule
    Method & Obj Geo. $\uparrow$ & Obj App. $\uparrow$ & Layout $\uparrow$ \\
    \midrule
    SimRecon  & $2.12 \pm 1.30$ & $1.86 \pm 1.25$ & $2.01 \pm 1.12$ \\
    SAM3D     & $2.37 \pm 1.40$ & $2.21 \pm 1.41$ & $2.26 \pm 1.22$ \\
    MV-SAM3D  & $2.66 \pm 1.59$ & $2.52 \pm 1.60$ & $2.68 \pm 1.01$ \\
    ShapeR    & $2.19 \pm 1.36$ & -               & $2.95 \pm 1.20$ \\
    Ours      & $\mathbf{3.81 \pm 1.25}$ & $\mathbf{3.78 \pm 1.33}$ & $\mathbf{4.08 \pm 0.95}$ \\
    \bottomrule
  \end{tabular}
\end{table}

\subsection{Settings}

\paragraph{Datasets}
We evaluate our method on 8 scenes from the synthetic Replica~\cite{replica19arxiv} dataset and 6 real-world scenes from ScanNet++~\cite{yeshwanthliu2023scannetpp}, following ShapeR~\cite{siddiqui2026shaper}. For each scene, we select every tenth view for testing and use the remaining views for training.

\paragraph{Metrics}
We adopt \ac{cd}, \fscore, and \ac{nc} to evaluate reconstruction quality, and PSNR, SSIM, and LPIPS for rendering. 
To further evaluate the perceptual quality of the reconstructed scenes, we conduct a user study to collect subjective scores from human subjects. Specifically, we focus on object geometry and appearance to evaluate the fidelity of individual entities, and spatial layout to assess whether the reconstructed results strictly adhere to the original scene.

\paragraph{Baselines}
We compare our framework against \sota decompositional reconstruction methods including SimRecon~\cite{xia2026simrecon}, ReconViaGen~\cite{chang2025reconviagen}, CUPID~\cite{huang2025cupid}, SAM3D~\cite{chen2025sam}, MV-SAM3D~\cite{li2026mv}, and ShapeR~\cite{siddiqui2026shaper}. For methods requiring point cloud inputs, \ie, SAM3D, MV-SAM3D, and ShapeR, we back-project depth estimated by Depth-Anything 3~\cite{lin2025depth} using the corresponding camera poses and instance masks to ensure a high-quality geometric initialization for fair comparison. For ReconViaGen and CUPID, we follow their official multi-view inference procedures and align the outputs to scene coordinates with FoundationPose~\cite{wen2024foundationpose}. Since ShapeR does not model appearance, we evaluate it only on geometry.

\subsection{Results}

\begin{figure*}[t]
  \includegraphics[width=\textwidth]{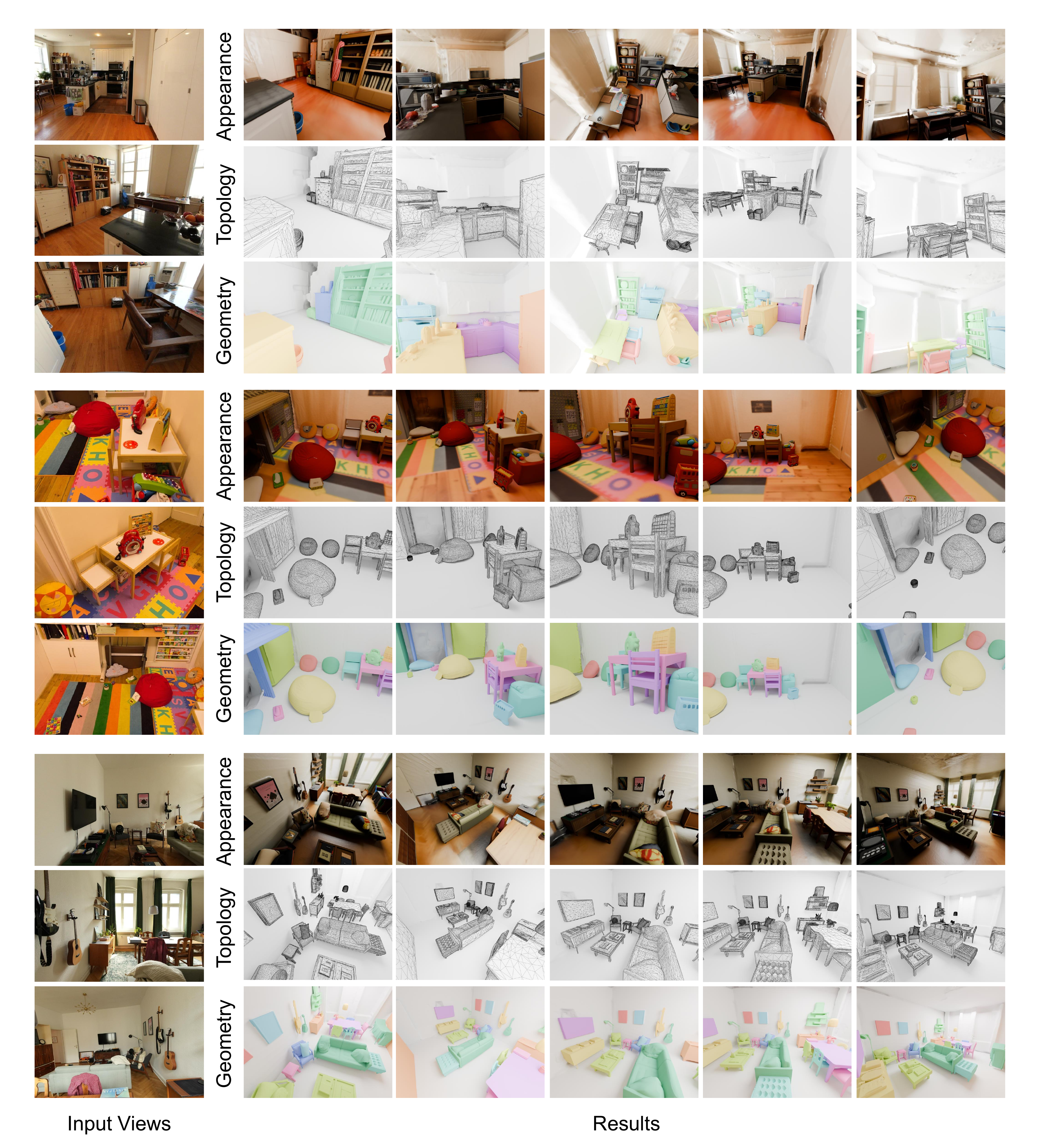}
  \caption{\textbf{Comprehensive Reconstruction Results.} Our method demonstrates robust, high-quality scene decompositional reconstruction performance on complex real-world captures.}
  \label{fig:qual_results_s2}
\end{figure*}

\begin{figure*}[t]
  \includegraphics[width=\linewidth]{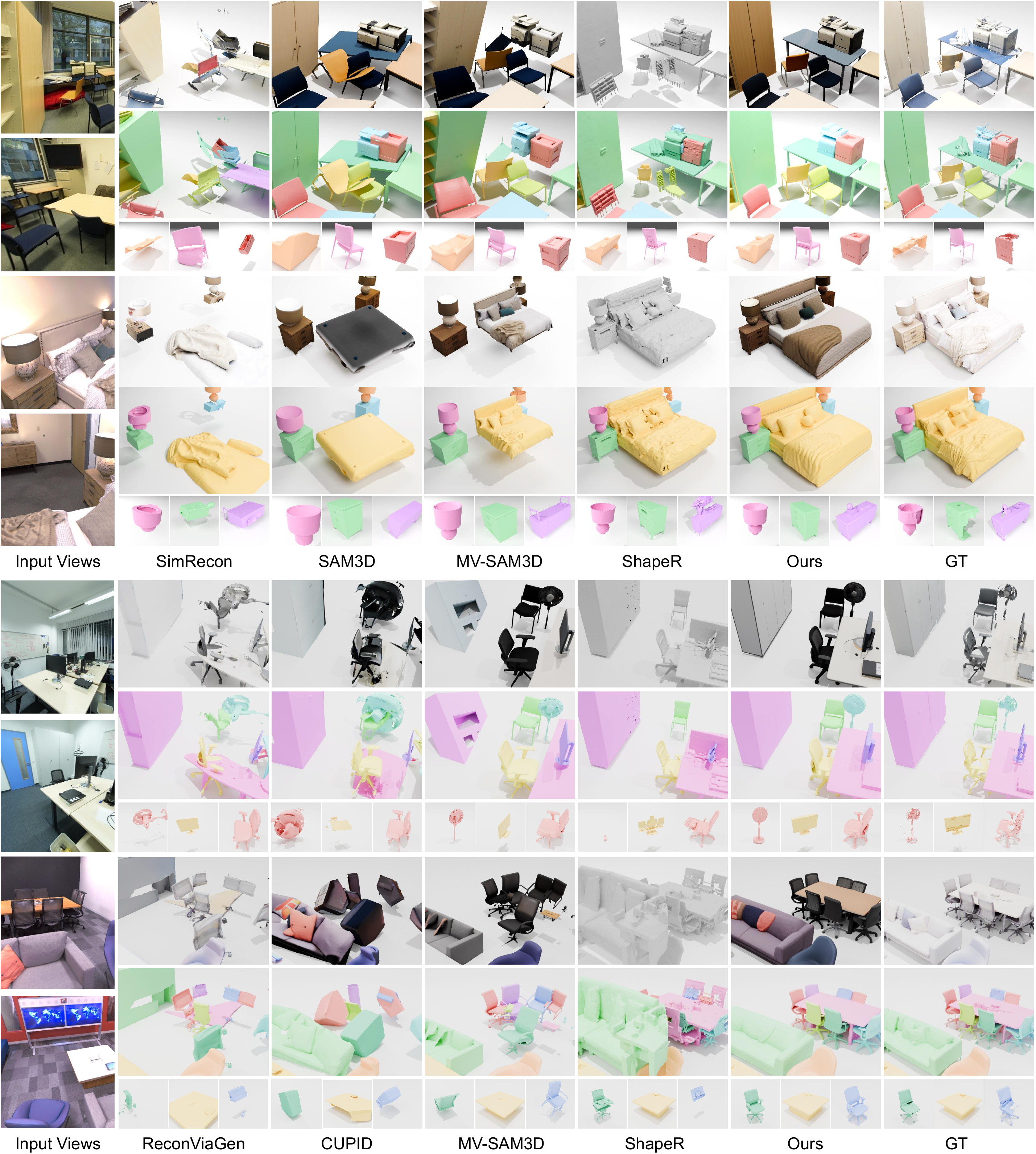}
  \caption{\textbf{More Qualitative Comparison.} Our method consistently outperforms other methods across different datasets.}
  \label{fig:qual_results_s1}
\end{figure*}

\paragraph{Geometry Reconstruction} As illustrated in~\cref{tab:main_results,fig:qual_results,fig:qual_results_s1,fig:qual_results_s2}, baselines that rely on decoupled 3D generation and post-hoc alignment, \eg, SimRecon, ReconViaGen, and CUPID, do not optimize objects directly in the original scene coordinates. Without explicit scene-level constraints, they are more susceptible to geometric distortion and accumulated alignment errors. Meanwhile, methods that utilize 3D point clouds as feed-forward conditions, \eg, SAM3D, MV-SAM3D, and ShapeR, frequently exhibit significant spatial drift in object position, rotation, and scale. These methods are also sensitive to input point-cloud density and may produce incomplete geometry on complex real-world datasets such as ScanNet++.
In contrast, our method achieves complete reconstruction while faithfully preserving the original spatial layout. By incorporating persistent explicit 3D guidance within the denoising optimization, we effectively anchor the stochasticity and inductive bias of the generative prior to the specific scene context. 
This advantage is further supported by our user study in~\cref{tab:user_study}, where human evaluators consistently favored our results for their high-quality reconstruction and precise spatial alignment.

\paragraph{Appearance Completion Quality}
Similar to our findings in geometry reconstruction, our guided denoising optimization significantly enhances appearance synthesis. Our method achieves style-consistent texture completion in unobserved regions while maintaining high consistency with original images in visible areas. Conversely, 3D-native baselines often produce textures that deviate obviously from the source style due to their lack of explicit appearance constraints, resulting in inferior rendering metrics. Qualitative assessments from human experts on texture consistency further validate that our optimization framework provides the most plausible appearance completion.

\begin{figure}[t]
    \centering
    \begin{subfigure}[b]{0.32\linewidth}
        \centering
        \includegraphics[width=\linewidth]{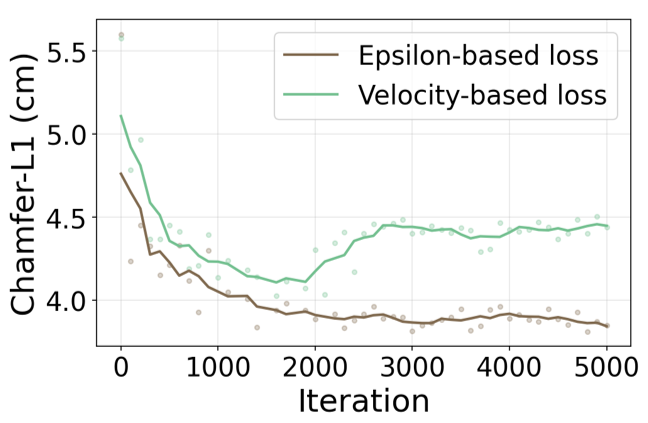}
        \caption{Denoising Loss Term}
        \label{fig:loss_func}
    \end{subfigure}
    \hfill
    \begin{subfigure}[b]{0.65\linewidth}
        \centering
        \includegraphics[width=\linewidth]{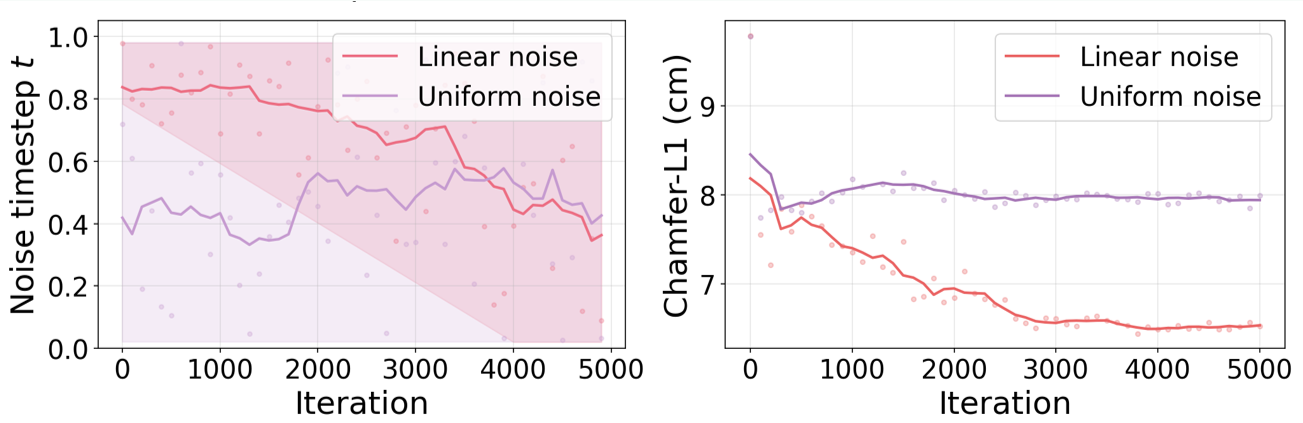}
        \caption{Noise Sampling $t$ Schedule}
        \label{fig:noise_schedule}
    \end{subfigure}

    \begin{subfigure}[b]{0.32\linewidth}
        \centering
        \includegraphics[width=\linewidth]{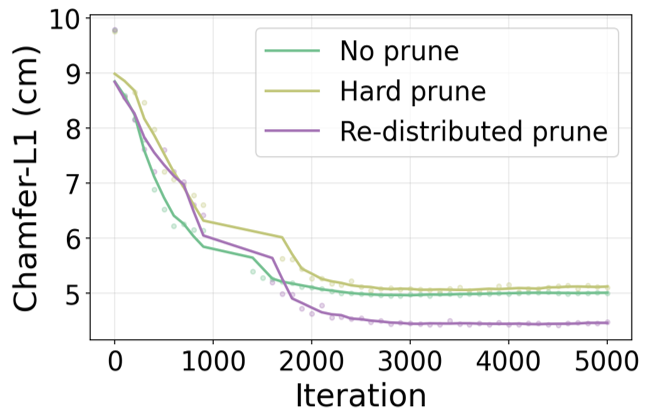}
        \caption{Pruning Strategy}
        \label{fig:prune}
    \end{subfigure}
    \hfill
    \begin{subfigure}[b]{0.65\linewidth}
        \centering
        \includegraphics[width=\linewidth]{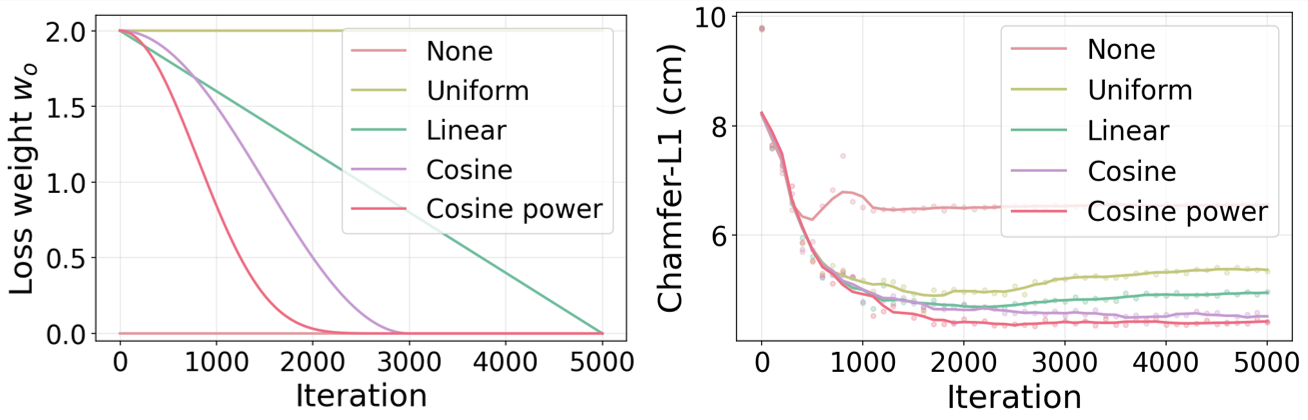}
        \caption{Preservation Loss Weight $w_O$ schedule}
        \label{fig:guidance_schedule}
    \end{subfigure}

    \caption{\textbf{Comparison of different optimization strategies.} Our epsilon-based denoising loss (a), linear noise schedule (b), re-distributed pruning strategy (c), and cosine-power preservation loss weight schedule (d) consistently achieve the most stable convergence.}
    \label{fig:strategy_results}
\end{figure}

\paragraph{Analysis of Design Choices}
We conduct an in-depth analysis of our key design components, as visualized in~\cref{fig:strategy_results}. First, we investigate the impact of the distillation loss formulation. As illustrated by the convergence curves in~\cref{fig:loss_func}, our proposed epsilon-based distillation loss facilitates a consistent and stable reduction in reconstruction error. In contrast, the naive velocity-based loss leads to significantly higher error in the later stages. This failure stems from an over-reliance on the flow matching model at high noise levels, where unstable optimization directions lead to sub-optimal local minima.
Furthermore, we validate the effectiveness of our temporal annealing strategies. As shown in~\cref{fig:noise_schedule}, employing our linear noise sampling schedule results in a more stable and monotonic improvement in reconstruction quality compared to fixed noise levels. Regarding the spatial guidance in vacant regions, we observe that simply zeroing out latent features in these regions (\ie, Hard Prune) yields negligible improvements over no pruning in~\cref{fig:prune}. Conversely, our re-distributed pruning strategy significantly enhances spatial fidelity by better aligning with the prior manifold. Finally, we examine the scheduling of the structural preservation loss weight $w_O$ in~\cref{fig:guidance_schedule}. While this regularization consistently boosts performance, adopting an appropriate annealing schedule (\eg, Cosine or Cosine-power) further achieves a better trade-off between generative flexibility and structural persistence, leading to the highest reconstruction fidelity.
Additionally, \cref{tab:efficiency} shows that our method trades longer runtime for better geometric completeness, texture consistency, and spatial alignment while maintaining comparable average memory.

\begin{table}[t]
\centering
\small
\caption{\textbf{Ablation study.} Geometry (GP) and appearance (AP) priors improve the reconstruction and rendering quality, while adaptive spatial guidance (G-SG and A-SG) further enhances the consistency.}
\label{tab:ablation_study}
\resizebox{\linewidth}{!}{
\begin{tabular}{cccc ccc ccc}
\toprule
\multirow{2}{*}{GP} & \multirow{2}{*}{G-SG} & \multirow{2}{*}{AP} & \multirow{2}{*}{A-SG} & \multicolumn{3}{c}{Reconstruction} & \multicolumn{3}{c}{Rendering} \\
\cmidrule(lr){5-7} \cmidrule(lr){8-10}
& & & & CD$\downarrow$ & F-score$\uparrow$ & NC$\uparrow$ & PSNR$\uparrow$ & SSIM$\uparrow$ & LPIPS$\downarrow$ \\
\midrule
$\times$ & $\times$ & $\times$ & $\times$ & 19.48 & 71.21 & 45.94 & 19.81 & 0.881 & 0.133 \\
\checkmark & $\times$ & $\times$ & $\times$ & 6.52 & 66.34 & 72.81 & 20.03 & 0.860 & 0.129 \\
\checkmark & \checkmark & $\times$ & $\times$ & \textbf{3.91} & \textbf{78.16} & \textbf{78.99} & 21.32 & 0.874 & 0.117 \\
\checkmark & \checkmark & \checkmark & $\times$ & 4.04 & 77.48 & 78.72 & 20.55 & 0.866 & 0.122 \\
\checkmark & \checkmark & \checkmark & \checkmark & 3.95 & 77.87 & 78.85 & \textbf{24.21} & \textbf{0.907} & \textbf{0.114} \\
\bottomrule
\end{tabular}
}
\end{table}

\subsection{Ablation Studies}
\label{sec:ablation}

Ablative studies, as shown in \cref{tab:ablation_study}, evaluate the contributions of the geometry prior (\textit{GP}), appearance prior (\textit{AP}), and their respective adaptive spatial guidance (\textit{G-SG} and \textit{A-SG}). Our analysis yields the following insights:
\begin{itemize}
    \item While incorporating the geometry prior (\textit{GP}) improves reconstruction completeness, its inherent stochasticity often introduces geometric deviations from the original scene. Furthermore, using the appearance prior (\textit{AP}) without guidance slightly degrades rendering performance, as unconstrained generation tends to conflict with the visual observation.

    \item Integrating adaptive spatial guidance (\textit{G-SG} and \textit{A-SG}) consistently yields performance gains. These results demonstrate that by enforcing deterministic spatial anchors and dynamic scheduling for noise sampling, the guidance mechanism effectively suppresses the generative prior's stochastic bias, ensuring both spatial and stylistic fidelity relative to the original scene.

    \item Our full model achieves superior performance across all metrics, validating the synergistic effect of the entire guided denoising optimization. This demonstrates that our framework successfully balances generative capacity with strict adherence to scene constraints, achieving complete and faithful scene reconstruction.
\end{itemize}

\begin{table}[t]
    \centering
    \small
    \caption{Runtime (hours) and GPU memory (GB) on Replica.}
    \label{tab:efficiency}
    \begin{tabular}{lccc}
        \toprule
        Method & Runtime$\downarrow$ & Peak Memory$\downarrow$ & Avg. Memory$\downarrow$ \\
        \midrule
        SimRecon    & 0.18 & 17.659 & 13.446 \\
        SAM3D       & \textbf{0.13} & 32.797 & 22.997 \\
        MV-SAM3D    & 3.13 & 68.157 & 51.580 \\
        ShapeR      & 0.37 & \textbf{11.255} & \textbf{5.691} \\
        ReconViaGen & 0.18 & 14.911 & 7.348 \\
        CUPID       & 0.14 & 13.778 & 7.123 \\
        Ours        & 1.35 & 20.965 & 7.450 \\
        \bottomrule
    \end{tabular}
\end{table}

% --------------------------- Section Separation Line --------------------------- 
\section{Conclusion}
In this work, we introduced \modelname, a framework that establishes a principled bridge between 3D-native generative priors and neural scene reconstruction through guided in-situ denoising optimization. By integrating a reformulated epsilon-based distillation loss with adaptive spatial guidance, our approach suppresses generative hallucinations and mitigates spatial drift. Extensive evaluations on Replica and ScanNet++ demonstrate that \modelname faithfully preserves structural fidelity and spatial layout, delivering textured meshes with clean topology, geometry, and appearance.
By providing a robust solution for recovering high-fidelity object entities within complex scenes, this framework has the potential to enhance the capacity to perceive and interact with the physical world in embodied AI and spatial intelligence.

\bibliographystyle{ACM-Reference-Format}
\bibliography{reference}

% --------------------------- Section Separation Line --------------------------- 

\clearpage

% --------------------------- Supplementary Material ---------------------------
\clearpage
\appendix
\section*{Supplementary Material}

This supplementary material provides technical and experimental details that complement the main paper. \Cref{sec:supp_derivation} first derives the epsilon-based distillation loss. \Cref{sec:supp_more_impl} then presents the overall pipeline, the implementation and scheduling of adaptive spatial guidance, and optimization parameters. Finally, \cref{sec:supp_eval,sec:supp_limitations} document the evaluation protocols and discuss limitations and future directions, respectively.

\section{Epsilon-Based Distillation for Flow Matching}
\label{sec:supp_derivation}

Building upon the analysis in the main text, the object geometry obtained from the initial reconstruction is treated as a latent representation $x'$ that requires refinement. Since the exact noise level of the initial reconstruction is unknown, we adopt an iterative denoising optimization framework to refine and complete the object geometry by leveraging the generative prior of a pre-trained Flow Matching model.
\paragraph{Loss Formulations} For a pre-trained flow model $v_\theta$, we define two possible formulations for optimization. The first, which we refer to as the epsilon-based refinement loss ($\mathcal{L}_{\epsilon}$), utilizes the predicted noise residual:
\begin{equation}
\label{eq:supp_epsilon_loss}
    \mathcal{L}_{\epsilon} = \mathbb{E}_{t, \epsilon} \left[  \| \hat{\epsilon}_\theta(x_t; t) - \epsilon \|_2^2 \right],
\end{equation}
where $x_t = (1-t)x' + t\epsilon$. Here, $\hat{\epsilon}_\theta(x_t; t) = x_t + (1-t)v_\theta(x_t, t)$ represents the noise estimate. 

Alternatively, a more direct formulation based on the velocity field is the Velocity-based refinement loss ($\mathcal{L}_{v}$):
\begin{equation}
\label{eq:supp_v_loss}
    \mathcal{L}_{v} = \mathbb{E}_{t, \epsilon} \left[\| v_\theta(x_t, t) - (\epsilon - x') \|_2^2 \right],
\end{equation}
where $(\epsilon - x')$ is the target vector field for the linear probability path.

\paragraph{Gradient Equivalence and Derivation} We formally show that these two formulations are equivalent in their gradient direction, differing only by a time-dependent scaling factor. Following the \ac{sds} paradigm~\cite{poole2022dreamfusion}, we consider the gradient of the loss with respect to the optimization variable $x'$ while omitting the computationally expensive and often unstable Jacobian of the model $\nabla_{x_t} v_\theta$. Starting from the noise residual relationship:
\begin{equation}
\begin{aligned}
\hat{\epsilon}_\theta - \epsilon &= [x_t + (1-t)v_\theta(x_t, t)] - \epsilon \\
&= [(1-t)x' + t\epsilon + (1-t)v_\theta(x_t, t)] - \epsilon \\
&= (1-t)x' + (t-1)\epsilon + (1-t)v_\theta(x_t, t) \\
&= (1-t) [v_\theta(x_t, t) - (\epsilon - x')].
\end{aligned}
\end{equation}
Now, consider the gradients of the two objectives with respect to $x'$. For the velocity-based loss, the simplified SDS gradient is:
\begin{equation}
\nabla_{x'} \mathcal{L}_{v} = \mathbb{E}_{t, \epsilon} \left[ \left( v_\theta(x_t, t) - (\epsilon - x') \right)  \frac{\partial x_t}{\partial x'} \right].
\end{equation}
Given that $\frac{\partial x_t}{\partial x'} = (1-t)$, we have:
\begin{equation}
\nabla_{x'} \mathcal{L}_{v} = \mathbb{E}_{t, \epsilon} \left[ (1-t) \left( v_\theta(x_t, t) - (\epsilon - x') \right) \right].
\end{equation}
For the epsilon-based loss $\mathcal{L}_{\epsilon}$, the gradient is:
\begin{equation}
\nabla_{x'} \mathcal{L}_{\epsilon} = \mathbb{E}_{t, \epsilon} \left[\left( \hat{\epsilon}_\theta - \epsilon \right) \frac{\partial x_t}{\partial x'} \right].
\end{equation}
Substituting the algebraic relationship $\hat{\epsilon}_\theta - \epsilon = (1-t) [v_\theta - (\epsilon - x')]$ into the above:
\begin{equation}
\nabla_{x'} \mathcal{L}_{\epsilon} = \mathbb{E}_{t, \epsilon} \left[(1-t)^2 \left( v_\theta(x_t, t) - (\epsilon - x') \right) \right].
\end{equation}
Comparing the two, we arrive at the relationship:
\begin{equation}
\nabla_{x'} \mathcal{L}_{\epsilon} = (1-t) \nabla_{x'} \mathcal{L}_{v}.
\end{equation}

\paragraph{Implications for Optimization} This derivation proves that $\mathcal{L}_{\epsilon}$ is a reweighted version of the velocity loss. In denoising tasks, it is desirable to provide conservative guidance when the noise level $t$ is high (preventing hallucinations) and precise guidance when $t$ is low (recovering fine details). The $(1-t)$ factor in the epsilon-based formulation naturally implements this weighting scheme by decaying the gradient magnitude as $t \to 1$. Empirically, this epsilon-based loss yields more robust convergence and superior geometric fidelity compared to the direct velocity-based alternative, as it inherently respects the reliability of the generative prior across different stages of the flow.

\section{More Implementation Details}
\label{sec:supp_more_impl}

In this section, we provide additional implementation details for the adaptive spatial guidance strategy, including spatial anchor construction and temporal scheduling, and then report the optimization parameters.

\subsection{Adaptive Spatial Guidance}
\label{sec:supp_anchors}
Adaptive spatial guidance combines occupied and vacant anchors with temporal scheduling. Occupied anchors preserve reliable observations, while vacant anchors suppress occupancy in invalid regions. Their guidance is progressively relaxed alongside the denoising timestep, enabling a transition from observation preservation to prior-driven refinement.

\paragraph{Occupied Anchors}
For each input voxel $i$ with edge length $s_i$ and octree level $l_i$, we compute its geometric uncertainty as
\begin{equation}
    U_{\mathrm{geom}}(i)=\frac{s_i}{\beta(l_i+l_0)},
    \qquad \beta=1,\quad l_0=1.
    \label{eq:supp_uncertainty}
\end{equation}
We min--max normalize the uncertainty values within each object and rasterize them into a grid $\widetilde{U}_{\mathrm{geom}}\in[0,1]^{64^3}$. When multiple input voxels cover the same cell, we retain the minimum uncertainty; uncovered cells are assigned an uncertainty of $1$. We then convert uncertainty to certainty and obtain the binary known-region mask using $\tau_C=0.03$:
\begin{equation}
    \begin{aligned}
        C_O(\mathbf v) &= 1-\widetilde{U}_{\mathrm{geom}}(\mathbf v), \\
        M_O(\mathbf v) &= \mathbf{1}\!\left[C_O(\mathbf v)\ge\tau_C\right].
    \end{aligned}
    \label{eq:supp_certainty_mask}
\end{equation}
This produces a soft certainty grid $C_O\in[0,1]^{64^3}$, where larger values indicate more reliable observations. The resulting occupied anchor is
\begin{equation}
    \Omega_O=\{\mathbf{v}\mid M_O(\mathbf{v})=1\}.
    \label{eq:supp_occupied_anchor}
\end{equation}
For geometry optimization, let $x'_{S,\mathrm{init}}$ denote the initial structural latent. The geometry preservation loss is
\begin{equation}
    \mathcal{L}_{O}^{\mathrm{geo}}
    =\frac{1}{|\Omega_O|}
      \sum_{\mathbf v\in\Omega_O}C_O(\mathbf v)
      \left\lVert x'_S(\mathbf v)-x'_{S,\mathrm{init}}(\mathbf v)\right\rVert_2^2.
    \label{eq:supp_geo_anchor_loss}
\end{equation}
The same occupied evidence guides appearance refinement while the initial sparse coordinates remain fixed. We smooth the certainty grid using a 3D Gaussian kernel with $\sigma=1$, sample it at the structured-latent coordinates, and square the sampled values. Let $c_n$ denote the squared certainty sampled at the coordinate of the $n$-th structured appearance feature, $N_L$ the total number of such features, and $x'_{L,\mathrm{init}}$ the initial appearance latent. The appearance preservation loss is
\begin{equation}
    \mathcal{L}_{O}^{\mathrm{app}}
    =\frac{1}{N_L}\sum_{n=1}^{N_L}c_n
      \left\lVert x'_L(n)-x'_{L,\mathrm{init}}(n)\right\rVert_2^2.
    \label{eq:supp_app_anchor_loss}
\end{equation}

\paragraph{Vacant Anchors}
Complementing $\Omega_O$, the vacant anchor $\Omega_V$ identifies regions in which the target object should not generate geometry. We construct this vacant-space evidence from two sources. First, the GeoSVR depth maps and input cameras define a scene-level visibility grid at resolution $R^3$ ($R=256$ in our experiments). A grid cell is retained as high-confidence vacant space if it lies in front of the observed surface in at least three views. Second, regions occupied by the background or another object are also invalid for the current object. We combine both sources in the structural grid and give occupied anchors precedence in the event of discretization overlap. The resulting high-confidence vacant set defines $\Omega_V$.

\paragraph{Re-distributed Prune}
As discussed in the main paper, directly setting the features in $\Omega_V$ to zero would disrupt the stochastic statistics learned by the prior. Our re-distributed pruning strategy instead shifts their distribution toward the void prior. Let $\mu$ be the current mean of $x'$ over $\Omega_V$, $\mu_0$ the target void-prior mean, and $\lambda_p(q)$ the scheduled redistribution factor at iteration $q$. We first compute
\begin{equation}
    \mu_{\mathrm{new}}(q)
    =\bigl(1-\lambda_p(q)\bigr)\mu+\lambda_p(q)\mu_0.
    \label{eq:supp_redistributed_mean}
\end{equation}
For each $\mathbf v\in\Omega_V$, we then independently resample the vacant-cell feature $x'(\mathbf v)$ from a uniform distribution with radius $r=2$:
\begin{equation}
    x'(\mathbf v)\sim
    \mathcal{U}\bigl(\mu_{\mathrm{new}}(q)-r,\mu_{\mathrm{new}}(q)+r\bigr).
    \label{eq:supp_redistribution}
\end{equation}
We set $\mu_0=\operatorname{sigmoid}(-7)=9.1\times10^{-4}\approx0.001$, corresponding to near-empty occupancy. We apply redistribution every 300 iterations during the early geometry stage and anneal its strength from $\lambda_p^0=0.2$ to zero over $N_V=2500$ iterations according to
\begin{equation}
    \lambda_p(q)=
    \begin{cases}
    \lambda_p^0\dfrac{1+\cos(\pi q/N_V)}{2},&q\le N_V,\\[4pt]
    0,&q>N_V.
    \end{cases}
    \label{eq:supp_prune_schedule}
\end{equation}
This operation discourages persistent occupancy in prohibited regions without collapsing all vacant features to a single value.

\paragraph{Temporal Scheduling}
In addition to defining where guidance is applied, we schedule both the denoising timestep and the occupied-anchor weight over optimization. During geometry refinement, we sample $t$ uniformly from a range whose lower bound decreases linearly from $0.8t_{\max}$ to $t_{\min}$ over the first 4000 iterations, while the upper bound remains fixed at $t_{\max}$. We use $t_{\min}=0.02$ and $t_{\max}=0.98$, so the sampling interval expands from approximately $[0.784,0.98]$ to $[0.02,0.98]$. Appearance refinement follows the same schedule over its first 2000 iterations.

We couple this timestep schedule with a cosine-power decay of the occupied-anchor weight. With $N_O=3000$ and initial weight $w_O^0=2$,
\begin{equation}
    w_O(q)=
    \begin{cases}
    w_O^0\left[\dfrac{1+\cos(\pi q/N_O)}{2}\right]^3,&q\le N_O,\\[4pt]
    0,&q>N_O.
    \end{cases}
    \label{eq:supp_occupied_schedule}
\end{equation}
The appearance-preservation weight similarly starts at $2$ and decays linearly over 3000 iterations. Overall, these schedules combine high-noise samples with strong spatial constraints early in optimization, then admit lower-noise samples as the constraints relax for prior-driven refinement.

\subsection{Optimization Parameters}
\label{sec:supp_impl}

\Cref{tab:supp_hyperparameters} lists the principal parameters that complement the schedules above. We use $N_G=5000$ iterations for geometry optimization and $N_A=3000$ iterations for appearance optimization. Both geometry and appearance variables are optimized with Adam and a cosine learning-rate decay. All pretrained TRELLIS modules~\cite{xiang2025structured} remain frozen throughout object refinement.

\begin{table}[t]
\caption{Primary optimization parameters.}
\label{tab:supp_hyperparameters}
\begin{lstlisting}[
    basicstyle=\ttfamily\footnotesize,
    frame=single,
    framesep=4pt,
    rulecolor=\color{black},
    columns=fullflexible,
    keepspaces=true,
    showstringspaces=false,
    breaklines=true,
    breakindent=1em,
    aboveskip=2pt,
    belowskip=0pt
]
geometry:
  grid_resolution: 64
  total_iters: 5000
  optimizer: Adam
  lr: 0.03
  lr_schedule: cosine
  timestep_range: [0.02, 0.98]
  timestep_anneal_iters: 4000
  distillation_weight: 1.0
  cfg_strength: 3.0
  cfg_interval: [0.5, 1.0]
  occupied_anchor_weight: 2.0
  occupied_anchor_schedule: cos_power
  occupied_anchor_end_iter: 3000
  occupied_certainty_threshold: 0.03
  redistribution_interval: 300
  redistribution_factor: 0.2
  redistribution_schedule: cosine
  void_logit: -7
  uniform_radius: 2.0

appearance:
  total_iters: 3000
  optimizer: Adam
  lr: 0.01
  lr_schedule: cosine
  timestep_range: [0.02, 0.98]
  timestep_anneal_iters: 2000
  cfg_strength: 3.0
  cfg_interval: [0.5, 1.0]
  preservation_weight: 2.0
  preservation_schedule: linear
  certainty_blur_sigma: 1.0
  certainty_power: 2.0
\end{lstlisting}
\end{table}

\section{Evaluation Details}
\label{sec:supp_eval}
This section specifies the datasets, metrics, baseline settings, and user-study procedure used in the main evaluation.

\subsection{Datasets}
We evaluate on eight Replica~\cite{replica19arxiv} scenes and six ScanNet++~\cite{yeshwanthliu2023scannetpp} scenes. Replica provides synthetic reference geometry and controlled renderings, while ScanNet++ contains real captures with incomplete and noisy observations. The same scene split, camera poses, and object masks are used for our method and all baselines. ScanNet++ uses the available dataset masks, which can be incomplete; the additional real-world qualitative examples use SAM2-generated masks~\cite{ravi2025sam}. These settings evaluate the method under moderate mask noise and occlusion without manually repairing every object mask.

\subsection{Reconstruction Metrics}
Following previous research~\cite{ni2024phyrecon}, we evaluate \ac{cd}, reported in $\mathrm{cm}$, \acs{fscore} at a threshold of $5\,\mathrm{cm}$, and \ac{nc} for 3D scene and object reconstruction. These metrics are defined in \cref{tab:metrics_definition}.

\begin{table}[ht!]
\caption{\textbf{Evaluation metrics.} We list the metrics used to evaluate reconstruction quality and their definitions. $P$ and $P^*$ are the point clouds sampled from the predicted and ground-truth meshes. $\bm{n}_{\bm{p}}$ is the normal vector at point $\bm{p}$.}
\small
\centering
    \setlength\extrarowheight{10pt}
    \begin{tabular}{lc}
    \toprule
    Metric & Definition \\
    \midrule
    \textbf{\acf{cd}}  & $\frac{\textit{Accuracy} + \text{Completeness}}{2} $ \\
    \textit{Accuracy} & $\underset{\bm{p} \in P}{\mbox{mean}}\left( \underset{\bm{p}^*\in P^*}{\mbox{min}} ||\bm{p}-\bm{p}^*||_1 \right)$ \\
    \textit{Completeness} & $\underset{\bm{p}^* \in P^*}{\mbox{mean}}\left( \underset{\bm{p} \in P}{\mbox{min}} ||\bm{p}-\bm{p}^*||_1 \right)$\\
    \midrule
    \textbf{F-score} & $\frac{ 2 \times \text{Precision} \times \text{Recall} }{\text{Precision} + \text{Recall}}$ \\
    \textit{Precision} & $\underset{\bm{p} \in P}{\mbox{mean}}\left( \underset{\bm{p}^*\in P^*}{\mbox{min}} ||\bm{p}-\bm{p}^*||_1 < 0.05 \right)$  \\
    \textit{Recall} & $\underset{\bm{p}^* \in P^*}{\mbox{mean}}\left( \underset{\bm{p} \in P}{\mbox{min}} ||\bm{p}-\bm{p}^*||_1 < 0.05\right)$ \\
    \midrule
    \textbf{Normal Consistency} & $ \frac{\textit{Normal Accuracy} + \textit{Normal Completeness}}{2} $\\
    \textit{Normal Accuracy}  &  $\underset{\bm{p} \in P}{\mbox{mean}}\left(  \bm{n}_{\bm{p}}^T\bm{n}_{\bm{p}^*}  \right) \,\, \text{s.t.} \, \, \bm{p}^* = \underset{\bm{p}^* \in P^*}{arg\,min} ||\bm{p}-\bm{p}^*||_1 $  \\
    \textit{Normal Completeness} &  $\underset{\bm{p}^* \in P^*}{\mbox{mean}}\left( \bm{n}_{\bm{p}}^T \bm{n}_{\bm{p}^*} \right) \,\, \text{s.t.} \, \, \bm{p} = \underset{\bm{p} \in P}{arg\,min} ||\bm{p}-\bm{p}^*||_1 $  \\
    \bottomrule
    \end{tabular}
\label{tab:metrics_definition}
\end{table}

\subsection{Rendering Metrics}
We render the composed meshes from held-out cameras and evaluate within the object/scene masks used by the benchmark. Images are converted to $[0,1]$. PSNR and SSIM use a unit data range. LPIPS uses the VGG backbone~\cite{simonyan2014very} and inputs scaled to $[-1,1]$. All methods are rendered with the same camera parameters and evaluation code.

\subsection{Baseline Protocol}
We use the official implementations and default model parameters for all baselines. For SimRecon~\cite{xia2026simrecon} and single-view SAM3D~\cite{chen2025sam}, we follow the unified view-selection protocol adopted by SimRecon. For SAM3D, MV-SAM3D~\cite{li2026mv}, and ShapeR~\cite{siddiqui2026shaper}, we back-project depth maps estimated by Depth-Anything 3~\cite{lin2025depth} within the same instance masks used by our method, transform the resulting points using the shared camera poses, and follow each method's official input point-density setting. 
For the object-centric generative baselines ReconViaGen~\cite{chang2025reconviagen} and CUPID~\cite{huang2025cupid}, which recover objects in method-specific coordinate systems, we follow their official multi-view inference procedures. Specifically, ReconViaGen recovers geometry in the VGGT~\cite{wang2025vggt} coordinate system, whereas CUPID reconstructs each object relative to its input camera coordinates. We then align their outputs with the scene coordinate system using FoundationPose~\cite{wen2024foundationpose}.

\subsection{User Study Protocol}
\label{sec:supp_user}
The user study contains 44 questionnaires completed by 44 unique expert raters. Each questionnaire contains 25 questions: 10 on object geometry, 10 on object appearance, and 5 on scene layout. Object questions are sampled uniformly from the objects in the eight Replica and six ScanNet++ scenes. Layout questions are sampled uniformly from the same 14 scenes.

The study is blind and randomized: method identities are hidden and presentation order is randomized. Object-geometry questions ask raters to assess geometric completeness and consistency with the visible evidence; object-appearance questions assess texture quality and consistency with observed views; scene-layout questions assess agreement of position, rotation, and scale with the input scene. Scores use a five-point scale, as reported in the main paper.

In addition to mean scores, we test whether our method is preferred to a competing baseline more often than chance. We use two-sided exact binomial tests with null probability $p=0.5$. The preferences are significant for scene layout, object geometry, and object appearance, with $p$-values $8.45\times10^{-75}$, $1.50\times10^{-58}$, and $4.26\times10^{-36}$, respectively. These tests support the preference trends in the rating table.

\section{Limitations and Future Work}
\label{sec:supp_limitations}
Our method relies on an initial GeoSVR reconstruction, object masks, and a pretrained 3D prior. Although our experiments on the real-world ScanNet++ scenes demonstrate robustness to incomplete masks and occlusion, severely inaccurate masks, poor initialization, or extreme occlusion may still leave insufficient reliable evidence for occupied anchors, while objects outside the prior distribution may not be faithfully completed.

Moreover, the current implementation uses the object-centric TRELLIS prior and has been evaluated primarily on indoor scenes with well-defined object instances. The in-situ formulation is not tied to a particular generative model and could therefore incorporate stronger and more diverse 3D priors. This offers a promising path toward larger or outdoor environments, such as street scenes, provided that reliable initial reconstruction and object decomposition are available. Regions that cannot be meaningfully decomposed into objects, such as sky or grassland, remain outside the current scope and may instead benefit from scene-level generative priors~\cite{wu2024reconfusion}.

A further practical limitation is computational cost. Per-object geometry and appearance optimization is slower than a single feed-forward pass, although multiple objects can be optimized in parallel. Future work could reduce latency by distilling the optimization process or developing more efficient guidance schedules while preserving reconstruction quality and spatial fidelity.

\end{document}